\documentclass[10pt,journal,compsoc]{IEEEtran}

\ifCLASSOPTIONcompsoc
  \usepackage[nocompress]{cite}
\else
  \usepackage{cite}
\fi

\usepackage{graphicx}
\usepackage{amsmath,amssymb} % define this before the line numbering.
\usepackage{color}

\usepackage{multirow}

\usepackage{amsmath}

\usepackage{wrapfig}
\usepackage{float}
\usepackage{breqn}

\usepackage{subfigure}
\usepackage{booktabs}

\usepackage[misc]{ifsym}

\begin{document}
\title{Deep Dense Multi-scale Network for Snow Removal Using Semantic and Geometric Priors }

\author{Kaihao Zhang,
        Rongqing Li,
        Yanjiang Yu,
        Wenhan Luo,
        Changsheng Li,
        and Hongdong Li
        \IEEEcompsocitemizethanks{\IEEEcompsocthanksitem Kaihao Zhang and Hongdong Li are with the College of Engineering and Computer Science, Australian National University, Canberra, ACT, Australia. E-mail: \{kaihao.zhang@anu.edu.au; hongdong.li@anu.edu.au\} \protect\\

\IEEEcompsocthanksitem Rongqi Li, Yanjing Yu and Changsheng Li are with the school of computer sciene and technology, Beijing Institute of Technology , Beijing, China. E-mail: \{lirongqing99@gmail.com; yuyanjiang87@gmail.com; lcs@bit.edu.cn\}  \protect\\

\IEEEcompsocthanksitem Wenhan Luo is with Tencent, Shenzhen, China. E-mail: \{whluo.china@gmail.com\}  \protect\\}

% \IEEEcompsocthanksitem Corresponding author: Changsheng Li} % <-this % stops a space
\thanks{Manuscript received April 19, 2005; revised August 26, 2015.}}


\markboth{Journal of \LaTeX\ Class Files,~Vol.~14, No.~8, August~2015}%
{Shell \MakeLowercase{\textit{et al.}}: Bare Advanced Demo of IEEEtran.cls for IEEE Computer Society Journals}



\IEEEtitleabstractindextext{%
\begin{abstract}

Images captured in snowy days suffer from noticeable degradation of scene visibility, which degenerates the performance of current vision-based intelligent systems. 
%Images captured in winter or at locations of high altitude often contain snow, which degenerates the performance of current vision-based intelligent systems. 
Removing snow from images thus is an important topic in computer vision. In this paper, we propose a Deep Dense Multi-Scale Network (\textbf{DDMSNet}) for snow removal by exploiting semantic and geometric priors. As images captured in outdoor often share similar scenes and their visibility varies with depth from camera, such semantic and geometric information provides a strong prior for snowy image restoration. We incorporate the semantic and geometric maps as input and learn the semantic-aware and geometry-aware representation to remove snow. In particular, we first create a coarse network to remove snow from the input images. Then, the coarsely desnowed images are fed into another network to obtain the semantic and geometric labels. Finally, we design a DDMSNet to learn semantic-aware and geometry-aware representation via a self-attention mechanism to produce the final clean images. Experiments evaluated on public synthetic and real-world snowy images verify the superiority of the proposed method, offering better results both quantitatively and qualitatively.

\end{abstract}

}

\maketitle

% is not selected <OR> if conference mode is selected - because compsoc
% conference papers position the abstract like regular (non-compsoc)
\IEEEdisplaynontitleabstractindextext

% \fi
\IEEEpeerreviewmaketitle

\section{Introduction}
\label{introduction}

\begin{figure*}[t] 
  \centering
  {\includegraphics[width=0.99\linewidth]{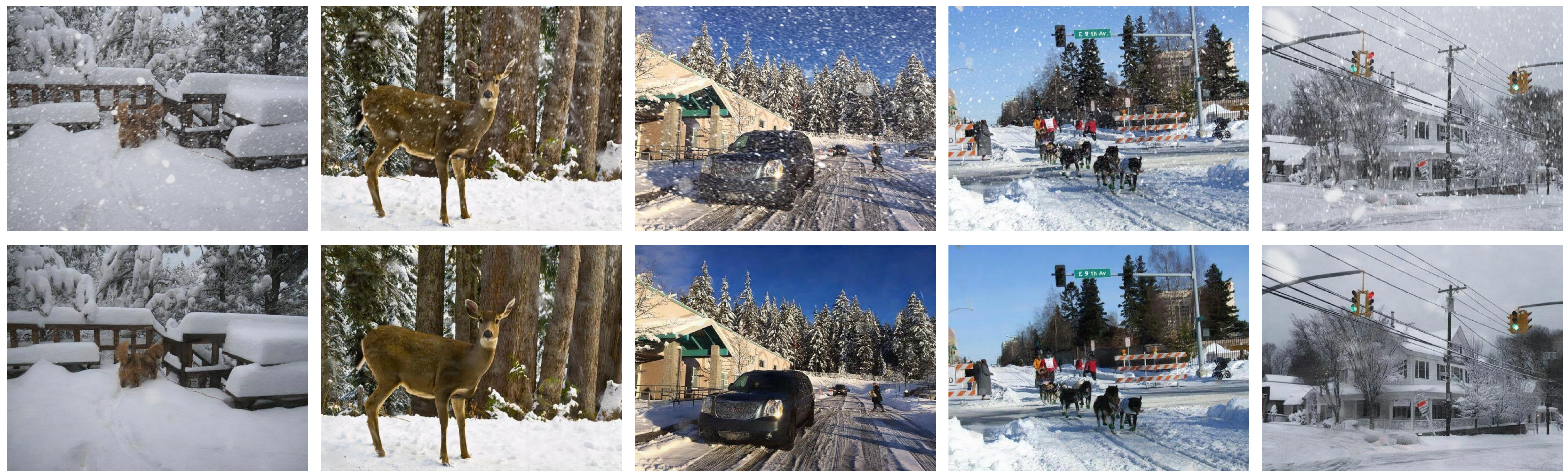}}
  \caption{{\bf Exemplar visual results of snow removal.} The goal of image desnowing is to remove snow and restore high-quality snow-free images. The various scene, occlusion and illumination variations of objects make it difficult to accomplish this task. Our proposed \textit{DDMSNet} is able to remove the snow and recover clean images with the aid of semantic and geometric priors.} 
  \label{fig:performance_introduction} 
\end{figure*}

As a common weather condition, snow can greatly affect the visibility of scene and objects in captured images. Its presence not only leads to poor visual quality but also degenerates the performance of subsequent image processing tasks, like object detection \cite{dalal2005histograms}, object tracking \cite{kaempfer2007three} and scene analysis \cite{itti1998model}. Therefore, snow removal from images is an important task and has attracted increasing attention in the computer vision community.

Compared with its counterpart task of image deraining, the snow removal is more challenging due to two reasons. Firstly, although the rain can also affect the visibility of objects in the scene, it is transparent which provides more scene information to remove rain and recover the high-quality clean images. On the contrary, the snow is opaque which makes it difficult for desnowing models to recover the occluded regions. 
Secondly, based on the definition of \cite{liu2018desnownet}, we consider only removing the snowflake in the air, rather than the snow falling on the ground or buildings. Because only the snowflake in the air can cover objects and has obviously negative impact on the vision-based intelligent systems. Therefore, it is important for snow removal models to understand semantic \cite{fourure2017residual} and geometry information.
%Secondly, in most cases, different from rain removal, we consider only removing the snow in the air, rather than the snow falling on the ground or buildings. Because only the snow in the air can cover objects and has obviously negative impact on the vision-based intelligent systems. Therefore, it is important for snow removal models to understand semantic \cite{fourure2017residual} and geometry information.

Existing single image desnowing methods adopt image priors \cite{wang2017hierarchical} or develop deep convolutional neural network (CNN) \cite{liu2018desnownet} to remove snow via learning a snow map or directly transferring the snowy images to their corresponding clean versions. Though the current state-of-the-art methods have achieved great success in snow removal, they ignore the geometric information which can affect the visibility of the same between the camera and objects, and the semantic information which can help to remove the snow in the air through understanding scenes. In addition, most of them focus on removing snow, but ignore to fill in the regions which are occluded by snow. Therefore, the recovered images fail to remove heavy snow and exhibit bad details.

To this end, we design a new deep CNN for snow removal via learning semantic-aware and geometry-aware representation. We leverage the semantic and geometric information as priors to train a deep learning based framework. The proposed framework consists of four sub-networks: a coarse snow removal network, a semantic segmentation network, a depth estimation network, and \textit{DDMSNet}. The snowy images are first fed into the coarse snow removal network to generate the 
coarsely desnowed images, which are then utilized to predict semantic segmentation and depth maps via a pre-trained semantic segmentation network and a depth estimation network. In order to generate finer results, a \textit{DDMSNet} is proposed. Different from the traditional multi-scale networks, the \textit{DDMSNet} applies dense connections between different scales in different dimensions, thus is able to achieve better performance to recover clean images with better details. In addition, \textit{DDMSNet} takes the coarsely desnowed images, maps of semantic segmentation and depth as input to calculate the semantic-aware and geometry-aware representation in an attention guided manner to restore clean images.

Moreover, considering that the snow always degrades the quality of images captured in the outdoor scenarios, which significantly deteriorates the performance of various core applications in autonomous driving, we create two new datasets specialized for street scenarios. These two datasets can be conveniently employed to evaluate algorithms of autonomous driving in the condition of snow. Along with another public dataset, the proposed method achieves the state-of-the-art performance for the task of snow removal.

Overall, the contributions of our method are summarized as follows.
\begin{itemize}
\item

Firstly, we propose a deep dense multi-scale network, named \textit{DDMSNet}, for single image desnowing. Different from previous multi-scale networks, which mainly capture features from pixel-level input, the proposed \textit{DDMSNet} can extract multi-scale representation from pixel-level and feature-level input.

%\textbf{Framework Level}: We construct a novel Enhanced Spatial-Temporal Interaction Network (\textit{ESTINet}) to remove rain streaks from videos. The proposed model firstly extracts spatial information and then learns the temporal correlations between extracted spatial features, rather than the input RGB images. Both of them interact each other to capture better spatial and temporal transformation. The final enhanced module further refines the deraining results, which helps the proposed model achieve better performance.

%Firstly, a multi-task shared learning deraining model, \textit{SADM}, is proposed to remove rain via scene understanding. This model not only considers pixel-level objective functions like previous methods, but also models the geometric structure and semantic information of input rainy images. Inside \textit{SADM}, a novel \textit{Semantic-Rethinking Loop} is employed to further strengthen the connection between scene understanding and image deraining. 
\item

Secondly, we exploit to use semantic and geometric information as priors for snow removal. Under a map-guided scheme, semantic and geometric features are obtained in different stages to help remove snow and recover clean images.

%\textbf{Backbone Level}: Among the proposed framework, we build a ResNet-based Encoder-Decoder backbone (\textit{SICM}), an Interaction-\textit{BCLSTM} backbone (\textit{STIM}) and a 3D-DenseNet backbone. Different from traditional \textit{BCLSTM}, our Interaction-BCLSTM architecture adds the direct features from last frame to the input and use convolutional operation to replace the \textit{tanh} function to adapt to different scales of input frames.

%Secondly, we propose \textit{PRRNet}, the first semantic-aware stereo deraining networks. The \textit{PRRNet} fuses the semantic information and multi-view information via \textit{SFNet} and \textit{VFNet}, respectively, to obtain the final stereo deraining images. %Meanwhile, the proposed method does not require semantic-annotated clean and rainy image pairs, which are expensive to collect.

\item

Thirdly, the two datasets we created will benefit research in the community. Experiments on three datasets show that the proposed method achieves the state-of-the-art performance on snow removal. Meanwhile, the desnowed images can improve the performance of many core applications, like semantic segmentation and depth estimation.

%\textbf{Experiment Level}: Experiments on three public rainy video datasets show that the proposed \textit{ESTINet} achieves the state-of-the-art performance of video deraining. Meanwhile, in terms of efficiency, via changing the numbers of input frames or light-weight backbones, the \textit{ESTINet} is faster than current state-of-the-art video deblurring methods with better performance.

%Thirdly, we synthesize two stereo rainy datasets for stereo deraining, which may be the largest datasets for stereo image deraining. Experiments on monocular and stereo rainy datasets show that the proposed \textit{PRRNet} achieves the state-of-the-art performance of both monocular and stereo deraining.
\end{itemize}

\section{Related Work}

Our work in this paper is closely related to snow removal, rain removal and haze removal, which are briefly introduced respectively in the following.

\subsection{Snow Removal}

Removing snow from a single image is a highly ill-posed problem. Its formulation can be described as,
\begin{equation}
O = A \odot M + B \odot (1 - M) \,,
\end{equation}
where $O$, $A$, $B$ and $M$ are the observed snowy image, the chromatic aberration map, the latent clean image and the snow mask, respectively.

Traditional methods utilize priors of snow-driven features to recover the clean images from the snowy versions. Bossu \textit{et al.} \cite{bossu2011rain} use a classical MoG model to separate the foreground from background. Then the snow can be detected from the foreground and removed to recover clean images under the help of HOG features. Similarly, Pei \textit{et al.} \cite{pei2014removing} extract features in color space and shape to detect the position of snow to help remove snow. In addition, Rajderkar \textit{et al.} \cite{rajderkar2013removing} and Xu \textit{et al.} \cite{xu2012improved,xu2012removing} employ frequency space separation and color assumptions to model the characteristics of snow for the snow removal task.

Recently, deep learning witnesses great success in low-level image enhancing tasks such as super super-resolution \cite{ledig2017photo,johnson2016perceptual}, image deblurring \cite{zhang2018adversarial,zhang2020deblurring,li2021arvo,zhang2020every}, image deraining \cite{liu2018desnownet,zhang2021dual,zhang2020beyond}, which also include snow removal. Specially, Liu \textit{et al.} \cite{liu2018desnownet} propose a DesnowNet, which is the first deep learning based method to remove snow from a single image. The DesnowNet adopts translucency and residual generation modules to recover image details obscured by snow. In order to generate realistic desnowed images, Li \textit{et al.} \cite{li2019single} adopt the GAN framework to restore better details. Li \textit{et al.} \cite{li2019stacked} introduce a multi-scale network for snow removal. However, they only consider the different scales in the pixel-level space, ignoring the feature space. More recently, Li \textit{et al.} \cite{li2020all} use the Network Architecture Search (NAS) framework to obtain a network, which achieves the state-of-the-art performance on snow removal. However, they ignore the semantic and geometric information, which are important priors for image restoration.

%\subsection{Semantic Understanding}
\subsection{Rain Removal}

Image deraining aims to remove the rain from rainy images and recover clean versions. The main difference from snow removal is that the snow is opaque and thus it is more difficult to restore the occluded details. In the recent decades, a set of methods are proposed to successfully remove rain via modeling the physical characteristics of rain \cite{kang2011automatic,luo2015removing,li2016rain,chang2017transformed,zhu2017joint,du2018single}. Kang \textit{et al.} \cite{kang2011automatic} propose a framework to first decompose the rainy images into low-frequency and high-frequency layers, and then use dictionary learning to remove rain in the high frequency layer. Chen \textit{et al.} \cite{chen2014visual} and Luo \cite{luo2015removing} use classified dictionary atoms and discriminative sparse coding to separate the rain and background. In \cite{li2016rain}, Li \textit{et al.} introduce a method to remove rain streaks via Gaussian mixture models.

It also has witnessed further promising achievement by the deep learning based deraining methods \cite{li2019single,zhang2019image,fu2017clearing,fu2017removing,yang2017deep,zhang2018density,li2018recurrent,eigen2013restoring,qian2018attentive,zheng2019residual}. Yang \textit{et al.} propose a deep CNN model for joint rain detection and removal. Fu \textit{et al.} \cite{fu2017removing} develop a deep detail network to remove rain and maintain texture details. However, it is difficult to remove heavy rain. Li \textit{et al.} \cite{li2019heavy} introduce a two-stage network to remove heavy rain. The first physics based stage decomposes the entangled rain streaks and rain accumulation, while the second model-free stage includes a conditional GAN to produce the final clean images.

%\subsection{Depth Estimation}
\subsection{Haze Removal}

Image dehazing \cite{schechner2001instant,shwartz2006blind,narasimhan2000chromatic,narasimhan2003contrast,nayar1999vision} aims to remove haze with other additional information such as atmospheric cues \cite{cozman1997depth,narasimhan2002vision} and depth information \cite{kopf2008deep,tarel2009fast}.

Early image dehazing methods rely on prior information to estimate the transmission maps and atmospheric light intensity. He \textit{et al.} \cite{he2010single} calculate a dark channel prior (DCP) based on the statistics of the outdoor images to help estimate the transmission map. Apart from the DCP-based methods \cite{tarel2009fast,meng2013efficient,li2015nighttime,nishino2012bayesian}, the attenuation prior is also utilized to recover clean images. Fattal \cite{fattal2014dehazing} derives a local formation model to explain the color-line in the context of hazy images and recover clean images. Berman \textit{et al.} \cite{berman2016non} introduce a novel non-local method based on the assumption that images can be present with a few hundreds of colors. Though these methods have shown their effectiveness in image dehazing, their performance in the real-world scene is not satisfied.

With the advance in deep learning methods, recent years have witnessed significant success in image dehazing \cite{cai2016dehazenet,ren2016single,zhang2018densely,li2017aod}. Many approaches develop different kinds of CNN models to recover clean images via estimating the transmissions and atmospheric light. Specially, Ren \textit{et al.} introduce a multi-scale dehazing network to remove haze with the coarse-to-fine scheme. Zhang and Patel \cite{zhang2018densely} and Li \textit{et al.} \cite{li2017aod} build a pyramid network and AOD-Net to estimate the transmission and atmospheric light for image restoration. Cheng \textit{et al.} \cite{cheng2018semantic} propose a deep semantic dehazing network which verifies the effectiveness of semantic information for image dehazing.

\begin{figure*}[t] 
  \centering
   \subfigure[The coarse snow removal network. D and U represent the down-sampling and up-sampling modules, respectively.]{
    \label{fig:overall_arc:a}
    \includegraphics[width= 0.99\linewidth]{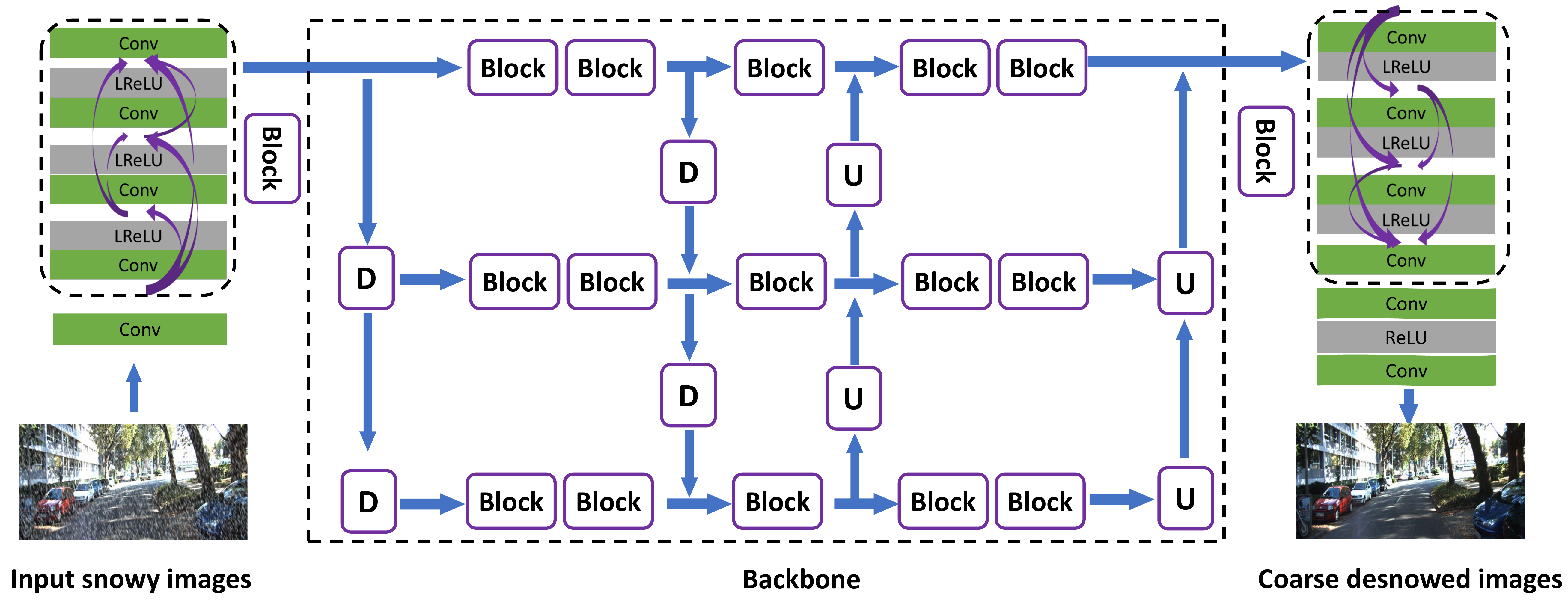}}
  \subfigure[Semantic and depth estimation]{
    \label{fig:overall_arc:b}
    \includegraphics[width=0.85\linewidth]{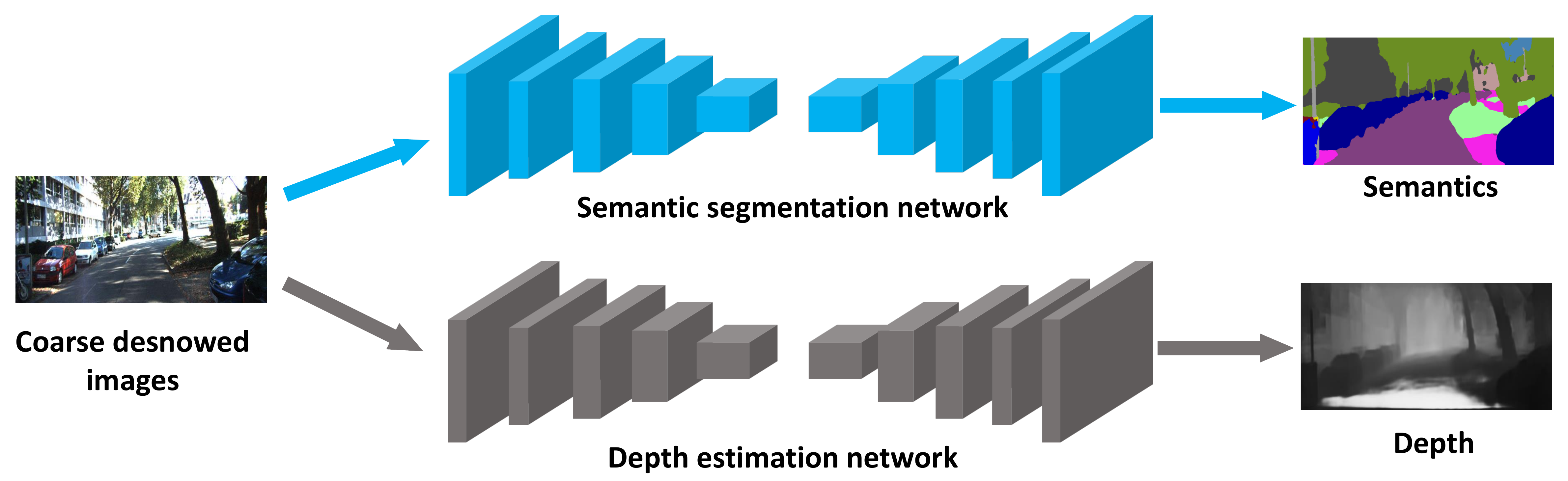}}
  \caption{{\bf The architecture of the proposed framework.} The coarse snow removal network consists of several Denseblocks \cite{huang2017densely}. The semantic segmentation and depth estimation networks are from \cite{tao2020hierarchical} and \cite{yin2019enforcing}, respectively.}
  \label{fig:overall_arc} 
\end{figure*}

\begin{figure*}[t] 
  \centering
  {\includegraphics[width=0.99\linewidth]{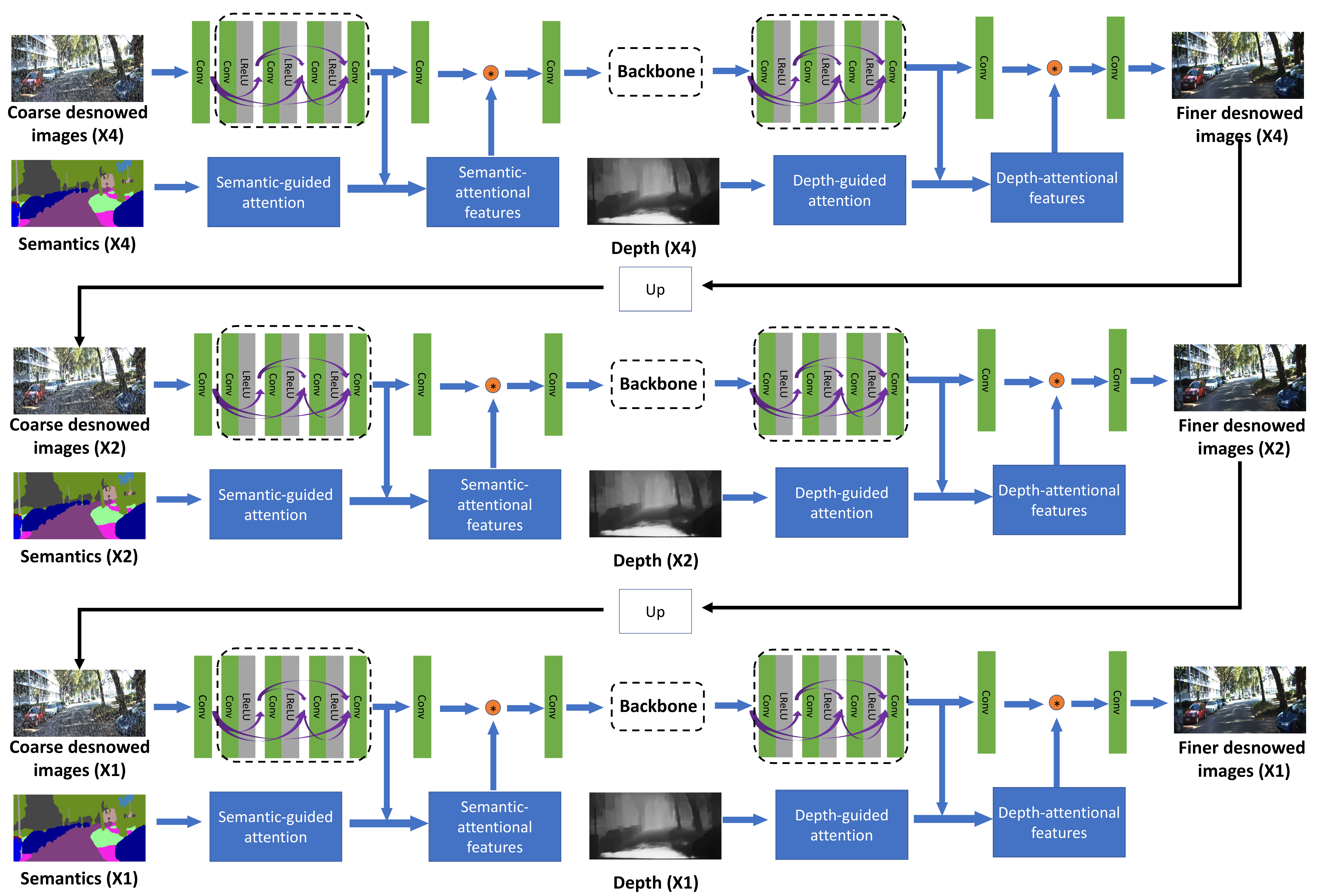}}
  \caption{{\bf The fine snow removal network.} The fine snow removal network is a dense multi-scale network, which extracts multi-scale features from both the RGB space and latent feature spaces. ``X" represents the scale of down-sampling. ``Up" means the up-sampling module.}
  \label{fig:overall_arc_2} 
\end{figure*}

\section{Our method}
\label{sec:method}

In this section, we first give an overview of the proposed snow removal framework. Then the details of the proposed methodology are introduced.

\subsection{Overall}
\label{sec:overall}

The goal of our work is to remove snow and recover clean images from their corresponding snowy versions. In order to improve the capability of restoration, we introduce the proposed \textit{DDMSNet} (Sec. \ref{sec:network}), which is able to extract multi-scale features from multiple dimensional representations. Considering that the semantic and geometric information can provide useful priors, we discuss in Sec. \ref{sec:semantic-aware} and \ref{sec:geometry-aware} how to obtain semantic-aware and geometry-aware representation using a map-guided manner to improve the performance. Finally, the loss functions are represent in Sec. \ref{sec:loss} to illustrate how to train the proposed framework. Specially, the overall of the framework is shown in Fig. \ref{fig:overall_arc}. The steps of the procedure are listed as follows.

\begin{itemize}
\item
A coarse image removal network is built to obtain coarsely desnowed images via reducing snow in the input images.

\item
A semantic segmentation and a depth estimation network are utilized to extract semantic labels and depth knowledge from the coarsely desnowed images.

\item
The coarsely desnowed images, semantic labels and depth maps are fed into the \textit{DDMSNet}. Based on a map-guided manner, semantic-aware and geometry-aware representation are obtained to help remove snow and obtain finer results. 

\end{itemize}

%\subsection{Densely Multi-Scale Network}
\subsection{Network Architecture}
\label{sec:network}

\textit{Coarse snow removal network.} We adopt a coarse-to-fine strategy to handle the task of snow removal. An input snowy image is addressed in a coarse stage and then processed by a fine stage with the semantic and geometry cues. To eliminate the possible negative effects of snow on the subsequent semantic understanding and depth estimation, we firstly in the coarse stage develop a coarse snow removal network to obtain pre-desnowed results, which can be represented as:
\begin{equation}
Y_{c} = G_{c}(O) \,,
\end{equation}
where $O$, $Y_{c}$ and $G_{c}$ are the observed snowy image, coarsely desnowed images and the coarse snow removal network, respectively. 

To be specific, the coarse snow removal network consists of three modules, \textit{i.e.,} a pre-processing module, a core module and a post-processing module. The pre-processing module includes one convolutional layer and a residual dense block. The channel number of the output feature map is 16. The core module consists of 5 dense blocks \cite{huang2017densely}. Each dense block contains 4 convolutional layers. Within each dense block, there are densely connected shortcuts among the convolutional layers. The number of feature maps is 16 in the sequential dense blocks. The feature maps in the same row have the same scale. \textit{D} represents the down-sampling module. Therefore, the scale of feature maps in the top row is double to the second row, whose scale is also double to the third row. Following the core module there is a post-processing module, which contains another dense block. This dense block includes 4 convolutional layers with ReLU as the activation function. In addition, there are another two convolutional layers in the post-processing module. The output is an image which is the coarse result of the snow removal.

%\fixme{you did not talk about the rows and columns in the core module. I guess you need to give more details.}

\textit{Deep dense multi-scale network (DDMSNet).} In order to obtain finer results, this paper introduces a new multi-scale network, which is shown in the Fig. \ref{fig:overall_arc_2}. The backbone in the Fig. \ref{fig:overall_arc_2} can refer to the Fig. \ref{fig:overall_arc}. We name it as deep dense multi-scale network because it not only extracts multi-scale features from multi-scale RGB images, but also extracts multi-scale features from single-scale RGB images. There are a set of sub-networks in the proposed DDMSNet, each sub-network corresponding to a scale. Specially, each sub-network consists of a semantic-guided attention module, a transfer module (which is the same as that in Fig. \ref{fig:overall_arc:a}) and a depth-guided attention module.

Roughly, in each sub-network of Fig. \ref{fig:overall_arc_2}, the semantic-aware module takes a RGB image and its corresponding semantic labels as input. With several convolutional layers and residual dense blocks (RDB) \cite{wang2018esrgan} in the semantic-aware module, feature maps are extracted, which are forwarded to the following transfer module to learn how to remove snow.
The transfer module is an attention-based multi-scale structure. It takes as input the feature maps generated by the semantic-attention module and extracts three-scale features. As the ``Backbone" in Fig. \ref{fig:overall_arc} shows, the rows represent different scales and the \textit{D} in each column are the connections between different scales. Each row consists of five RDB structures and the number of feature maps is fixed during the processing. The columns consist of down-sample and up-sample modules. The down-sample module consists of two convolutional layers to reduce the feature map to its half, while the up-sample module also includes two convolutional layers to increase the number of the feature maps by a factor of $2$, whose details can refer to the Fig. \ref{fig:overall_arc}. In order to concatenate the features from different scales, we adopt a channel attention manner,
\begin{equation}
F_a = \alpha_r F_r + \beta_c F_c \,,
\end{equation}
where $F_r$ and $F_c$ are the features from the row and column, respectively. $\alpha_r$ and $\beta_c$ are weights to balance different features.
Finally, the geometry-aware module takes the features from the transfer module and geometry labels as input to recover final snow-free images. To be specific, the geometry module consists of 3 layers of convolutional operation, and 3 residual dense blocks. The number of output feature channel is three, corresponding to an RGB image. 

%gridnet

\subsection{Semantic-aware Representation}
\label{sec:semantic-aware}
%ziang chen
%depth

\begin{figure*}[t] 
  \centering
  {\includegraphics[width=0.8\linewidth]{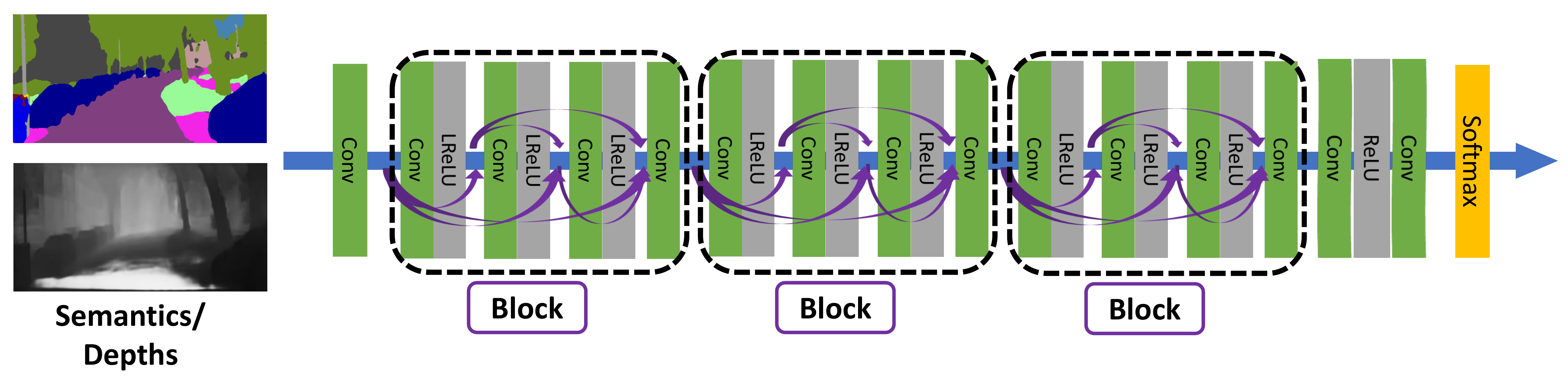}}
  \caption{{\bf The Architecture of network to learn the semantic and geometry attention weights}.}
  \label{fig:guid} 
\end{figure*}

In this section, we propose a method to explore the semantic information to help remove snow. We first use a pretrained semantic segmentation network \cite{tao2020hierarchical} to predict semantic labels based on the coarsely desnowed images, and then learn a semantic-aware representation under the guidance of semantic labels. Fig. \ref{fig:guid} shows the architecture of the mechanism. It processes the input features and semantic labels. The output is input into a Softmax function to transfer a set of attention weights ${A_1, A_2, ..., A_n}$ to ${W_1, W_2, ..., W_n}$. Each of them corresponds to a type of objects,

\begin{equation}
\label{semantic}
W_i =  \frac{e^{A_i}}{\sum_{c=1}^{n} e^{A_c}} \,,
\end{equation}
where $c$ is the channel of features.

The feature map before the semantic-aware representation has $30$ channels. We set $n$ as 30 because the semantic label set has about $30$ types of different objects. Therefore, feature maps are divided into $30$ groups. The semantic-aware representation is obtained based on $W$ and feature channels in an element-wise manner. After that, we use group convolution in the $30$ groups to process the semantic-aware representation and merge all the features from different groups via a $1 \times 1$ CNN layer.

\subsection{Geometry-aware Representation}
\label{sec:geometry-aware}

%depth

Similar to the method of generating semantic-aware representation, we first use a depth estimation network \cite{yin2019enforcing} to obtain depth information based on the coarsely desnowed images. Then the depth information is combined with features extracted from input images to obtain the geometry-aware representation. However, it is different from the above part of semantic-aware representation in terms of two aspects. Firstly, we concatenate the geometry information in the last layer of \textit{DDMSNet}. The input features are the output of the transfer module, rather than the low-level features. Secondly, the group number in Eq. \ref{semantic} is set to $8$, rather than $30$. Finally, the geometry-aware representation is input into two CNN layers to recover clean images.

\subsection{Loss Function}
\label{sec:loss}
%ziyi shen

We train the coarse snow removal network and the \textit{DDMSNet} with a smooth $L_1$ and the perceptual loss function. For the coarse desnowing network, the smooth $L_1$ loss function can be represented as:
\begin{equation}
\mathcal{L}_1 =  \frac{1}{N} \sum_{x=1}^{N} \sum_{i=1}^{3} Q (I'_i(x) - I_i(x)) \,,
\end{equation}
where
\begin{equation}
\left\{
             \begin{array}{lr}
             Q(e)=0.5e^2,  \ \ \ \ \ \ \ if \ |e|<1,  \\
             Q(e)=|e|-0.5,  \ \ otherwise.   &  
             \end{array}
\right.
\end{equation}
$I'_i(x)$ and $I_i(x)$ are the intensity of the $i$-th color channel of pixel $x$ in the desnowed image and the ground-truth image, respectively. $N$ is the batch size.

\begin{figure*}[!tb]
  \centering
  \subfigure[Input]{
    \label{ablation:a}
    \includegraphics[width=0.24\linewidth ]{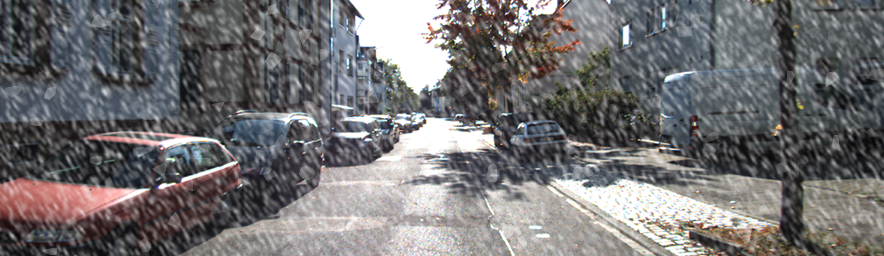}}
    \subfigure[SnowCNN]{
    \label{ablation:b}
    \includegraphics[width=0.24\linewidth ]{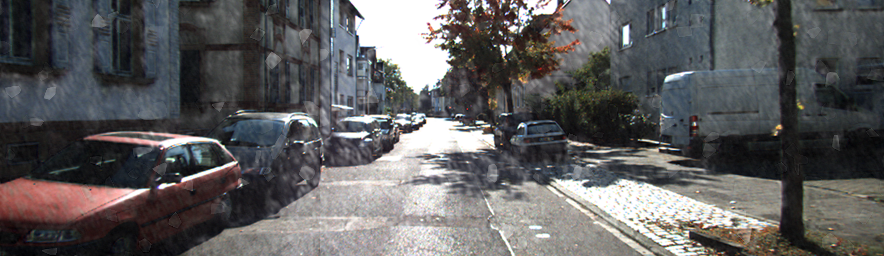}}
    \subfigure[MSNet]{
    \label{ablation:c}
    \includegraphics[width=0.24\linewidth ]{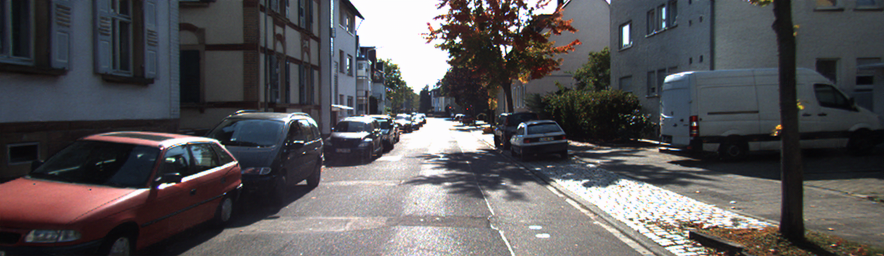}}
    \subfigure[DDMSNet]{
    \label{ablation:d}
    \includegraphics[width=0.24\linewidth ]{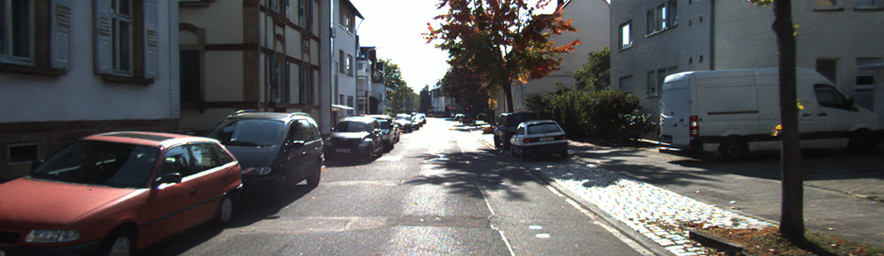}}
    \subfigure[DDMSNet (+)]{
    \label{ablation:d}
    \includegraphics[width=0.24\linewidth ]{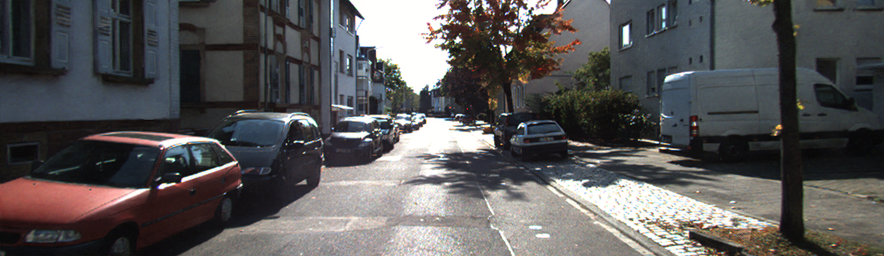}}
    \subfigure[DDMSNet (S)]{
    \label{ablation:e}
    \includegraphics[width=0.24\linewidth ]{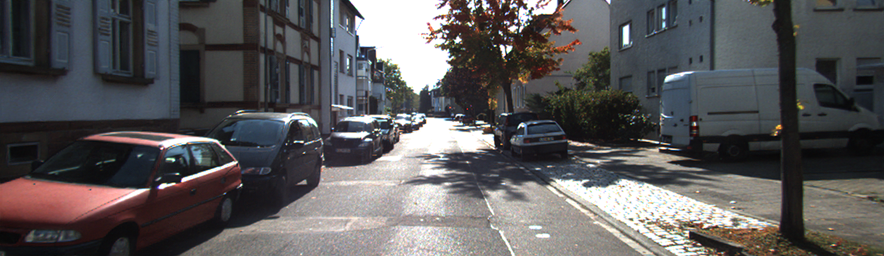}}
    \subfigure[DDMSNet(G)]{
    \label{ablation:e}
    \includegraphics[width=0.24\linewidth ]{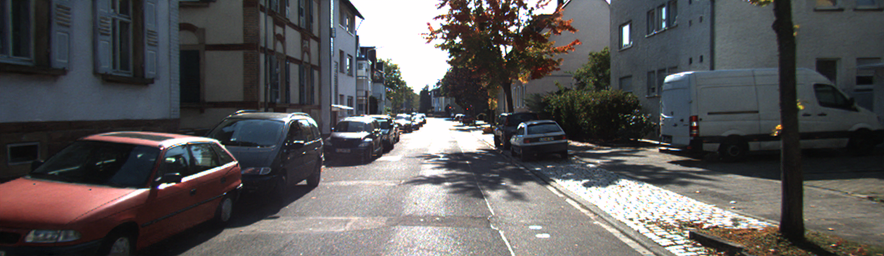}}
  \subfigure[DDMSNet(S+G)]{
    \label{ablation:f}
    \includegraphics[width=0.24\linewidth ]{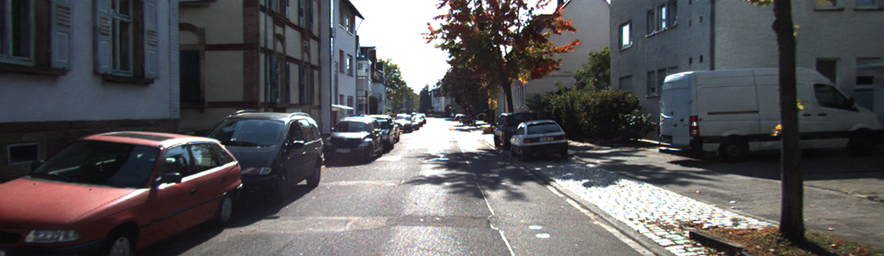}}
\caption{ {\bf Exemplar results on the SnowKITTI2012 dataset.} From top to bottom: input, SnowCNN, MSNet, DDMSNet, DDMSNet(+), DDMSNet(S), DDMSNet(G) and DDMSNet(S+G). Best viewed in color.}
  \label{fig:ablation} 
\end{figure*}

The perceptual loss function is defined as:
\begin{equation}
{\mathcal{L}_{p}} = \frac{1}{{CWH}}\sum\limits_{c = 1}^C \sum\limits_{x = 1}^W {\sum\limits_{y = 1}^H {{{(G(I^{clean})_{x,y,c} - G(I^{de})_{x,y,c})}^2}} }\, ,
\end{equation}
where $C$, $W$ and $H$ are the channel, width and height of feature maps extracted from a pre-trained VGG16 \cite{simonyan2014very} model. $I_{x,y,c}^{clean}$ is the pixel value of clean images at location $\left(x, y, c\right)$, and $G(I^{de})_{x,y,c}$ corresponds to the value of desnowed images.

The loss functions of the \textit{DDSMNet} are similar to those of the coarse desnowing network. The main differences come from the multi-scale scheme. Therefore, the smooth $L_1$ and perceptual loss functions for \textit{DDSMNet} are,
\begin{equation}
\mathcal{L}_1^{f} =  \frac{1}{M} \sum_{m=1}^{M} \mathcal{L}_1^{m} \,,
\end{equation}
\begin{equation}
\mathcal{L}_p^{f} =  \frac{1}{M} \sum_{m=1}^{M} \mathcal{L}_p^{m} \,,
\end{equation}
where $M$ is the number of different scales we adopt. We set it as $3$ in our paper. Here the superscript $f$ indicates the loss functions are for the fine snow removal network.

The total loss function is defined by combining two different loss functions as follows, 
\begin{equation}
\mathcal{L} = \mathcal{L}_1 + \beta * \mathcal{L}_p \,,
\end{equation}
where $\beta$ is set to $0.05$ in our paper. Note that, the formulation of loss function applies to both the coarse and the fine snow removal networks.

\begin{table*}
  \centering 
  %\small
    \caption{Performance comparison of different architectures on the SnowKITTI2012 dataset, in terms of PSNR and SSIM. Here ``Small", ``Medium" and ``Large" indicate the particle size of the snow.}
    \begin{tabular}{l | c c c }
    \toprule
    Methods &  Small & Medium & Large \\
    \hline %\midrule
    \textit{SnowCNN} & 34.94/0.9724 & 29.83/0.9396 & 31.14/0.9323 \\
    \textit{MSNet} & 36.17/0.9792 & 32.61/0.9586 & 32.70/0.9473 \\
    \hline
    \textit{DDMSNet}  & 37.99/0.9840  & 34.24/0.9677 & 34.12/0.9572 \\
    \textit{DDMSNet (+)}  & 37.23/0.9841 & 34.66/0.9720 & 34.16/0.9642 \\
    \textit{DDMSNet(G)}  & 38.15/0.9871 & 34.54/0.9736 & 34.94/0.9691 \\
    \textit{DDMSNet(S)}  & 38.89/0.9864 & 35.41/0.9740 & 35.22/0.9678 \\
    \hline
    \textit{DDMSNet(G+S)}  & \textbf{39.53/0.9877} & \textbf{35.50/0.9745} & \textbf{35.55/0.9700} \\
    \bottomrule
    \end{tabular}%
    \label{table:ablation}
\end{table*}%

\begin{figure*}[t] 
  \centering
  {\includegraphics[width=0.99\linewidth]{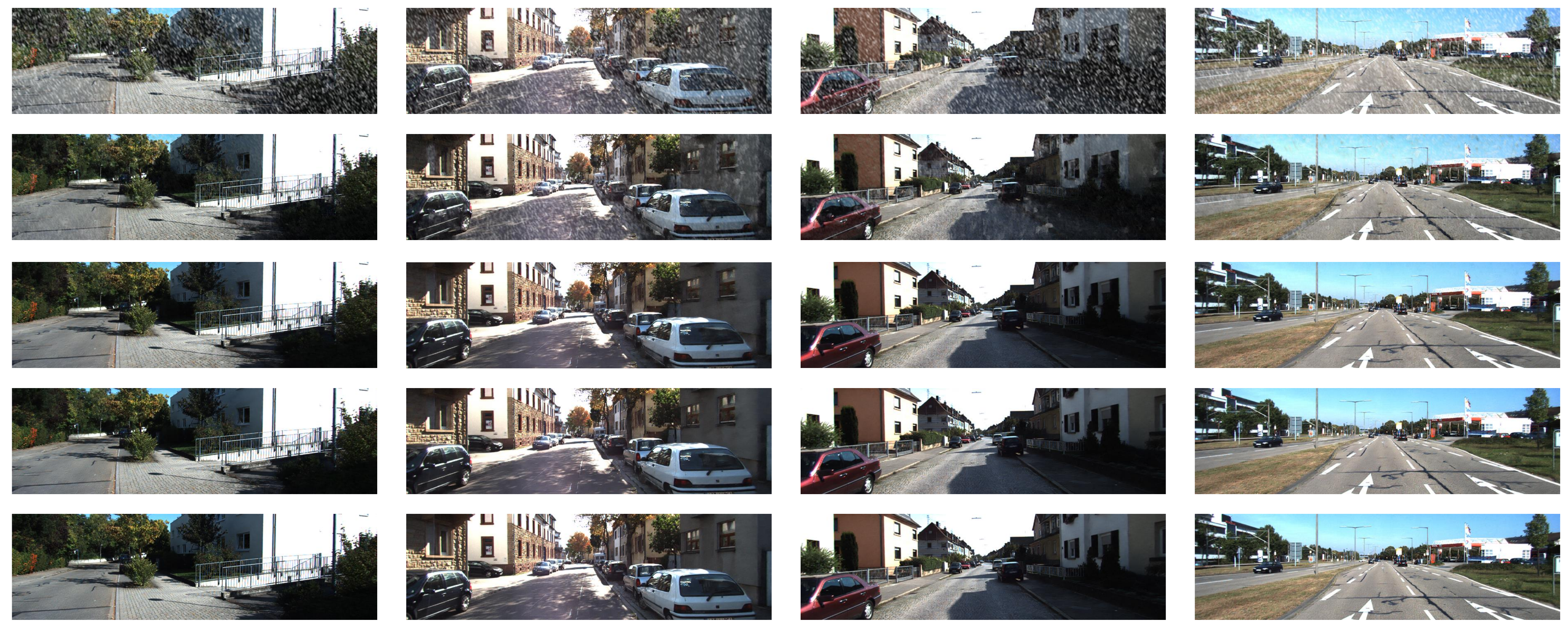}}
  \caption{{\bf Exemplar results on the SnowKITTI2012 dataset}. From top to bottom: input, DesnowNet, RESCAN, SPANet, and Ours. Best viewed in color.}
  \label{fig:kitti} 
\end{figure*}

\section{Experiments}

In this section, we evaluate our proposed method on three datasets of snowy images. Considering that the existing datasets such as the Snow100K dataset \cite{liu2018desnownet} lack snowy images of street scenes, we firstly create two new datasets of snowy images of street scenes, which are introduced in Sec. \ref{dataset}. Then the implementation details of our framework are introduced in Sec. \ref{implementation}. We conduct an ablation study to show the effectiveness of different modules in Sec. \ref{ablation}. Finally, we compare the proposed method with the state-of-the-art methods on both synthesized snowy images and real-world images to demonstrate its superiority in Sec. \ref{comparison}.

\subsection{Datasets}
\label{dataset}

\textbf{SnowKITTI2012 dataset.} We first use Photoshop to create a synthetic SnowKITTI2012 dataset based on the public KITTI 2012 dataset \cite{geiger2013vision}. The training and testing sets of the proposed dataset include $1,500$ and $1,000$ pairs of images, respectively. In order to model different types of snow, each set contains three kinds of snow including light, medium and heavy snow. The size of images in both the training and the testing sets is $884 \times 256$.

\textbf{SnowCityScapes dataset.} The SnowCityScapes dataset is created based on the Cityscapes dataset \cite{cordts2016cityscapes}. The training and testing sets consist of $2,000$ and $2,000$ pairs of images, respectively. The size of images in both the training and the testing sets is $512 \times 256$. Similar to the SnowKITTI2012 dataset, we also provides three kinds of snowy images.

\textbf{Snow100K dataset.} This dataset is created by Liu \textit{et al.} \cite{liu2018desnownet}. In order to model snowy images, they first produce $5,800$ snowy masks and download $100K$ clean images. Then snowy images are synthesized based on the clean images and snowy masks. This dataset provides three kinds of snow, \textit{i.e.}, small, medium, and large particle sizes. They also provide $1,329$ realistic snowy images to evaluate models in terms of generalization in the real world.

\subsection{Implementation Details}
\label{implementation}

In this paper, we use a Gaussian distribution with zero mean and a standard deviation of $0.01$ to initialize the parameters of our proposed networks. The size of mini-batch during the training stage is set to $8$ for updating the models. In order to boost the variance of the data, we augment data by cropping $224 \times 224$ patches from images at random locations, and randomly flipping them along the horizontal direction. The learning rate is set as $10^{-4}$ and then we decrease it to $10^{-6}$ after the training loss achieves convergence.

%The coarse desnowing network consists of a convolutional layer, six RRDB blocks and another convolutional layers. The size of feature maps is the same as the size of input images. For the finer desnowing network, \textit{DDMSNet}, the semantic-aware and geometry-aware modules include a convolutional layer, a RRDB block, another two convolutions layers. While the transfer model consists of five RRDB blocks. All the kernels are set to $3 \times 3$, except the last two convolutional layers in semantic-aware and geometry-aware layers, whose size is set to $1 \times 1$. \fixme{this paragraph seems unnecessary if architecture is introduced in the methodology part.}

\subsection{Ablation Study}
\label{ablation}

%\begin{figure}[t] 
%  \centering
%  {\includegraphics[width=0.99\linewidth]{Figures/ablation.pdf}}
%  \caption{{\bf Exemplar results on the SnowKITTI2012 dataset.}. From top to bottom: input, SnowCNN, MSNet, DDMSNet, DDMSNet (+), DDMSNet (S), DDMSNet(G), DDMSNet(S+G) and ground-truth. Best viewed in color.}
%  \label{fig:ablation} 
%\end{figure}

\begin{table*}[t]
  \centering 
  %\small
    \caption{Performance comparison with state-of-the-art methods on the SnowKITTI2012, SnowCityScapes and Snow100K datasets.}
    \begin{tabular}{l | c | c c c c}
    \toprule
    Dataset &  Snow & RESCAN & SPANet & DesnowNet  & Ours \\
    \hline %\midrule
     & Small & 35.68/0.9735 & 35.90/0.9781 & 32.11/0.9423 & \textbf{39.53/0.9877} \\
    \textit{SnowKITTI2012}  & Medium & 31.81/0.9489 & 32.08/0.9559 & 29.11/0.8903 & \textbf{35.50/0.9740} \\
      & Large & 32.33/0.9360 & 32.49/0.9468 & 29.14/0.8663 & \textbf{35.55/0.9700} \\
    \hline
      & Small & 38.59/0.9815 & 39.66/0.9872 & 35.39/0.9603 &  \textbf{42.24/0.9913} \\
    \textit{SnowCityScapes}  & Medium & 33.63/0.9627 & 35.73/0.9741 & 33.58/0.9382& \textbf{38.30/0.9826} \\
      & Large & 34.24/0.9624 & 35.50/0.9669 & 33.39/0.9148 & \textbf{38.60/0.9822} \\
    \hline
      & Small &31.51/0.9032 & 29.92/0.8260 & 32.33/0.9500 & \textbf{34.34/0.9445} \\
    \textit{Snow100K}  & Medium & 29.95/0.8860 & 28.06/0.8680 & 30.87/0.9409 & \textbf{32.89/0.9330}\\
      & Large & 26.08/0.8108 & 23.70/0.7930 & 27.17/0.8983 & \textbf{28.85/0.8772} \\
    \bottomrule
    \end{tabular}%
    \label{table:sota}
\end{table*}%

The proposed \textit{DDMSNet} has the advantage of extracting dense multi-scale features from the input images. Meanwhile, the semantic-aware and geometry-aware representations provide semantic and geometric priors to help update the \textit{DDMSNet} to learn how to remove snow and restore clean images. In order to verify their effectiveness, we perform an ablation study by evaluating seven variant networks: \textit{SnowCNN}, \textit{MSNet}, \textit{DDMSNet}, \textit{DDMSNet (+)}, \textit{DDMSNet (S)}, \textit{DDMSNet(G)} and \textit{DDMSNet(S+G)}.

\begin{itemize}
\item
\textbf{SnowCNN} is a plain CNN network consisting of one convolutional layer, $7$ RRDB and another convolutional layer. The input of this model is a pair of (clean and snowy) images of original size without scaling.

\begin{figure*}[t] 
  \centering
  {\includegraphics[width=0.99\linewidth]{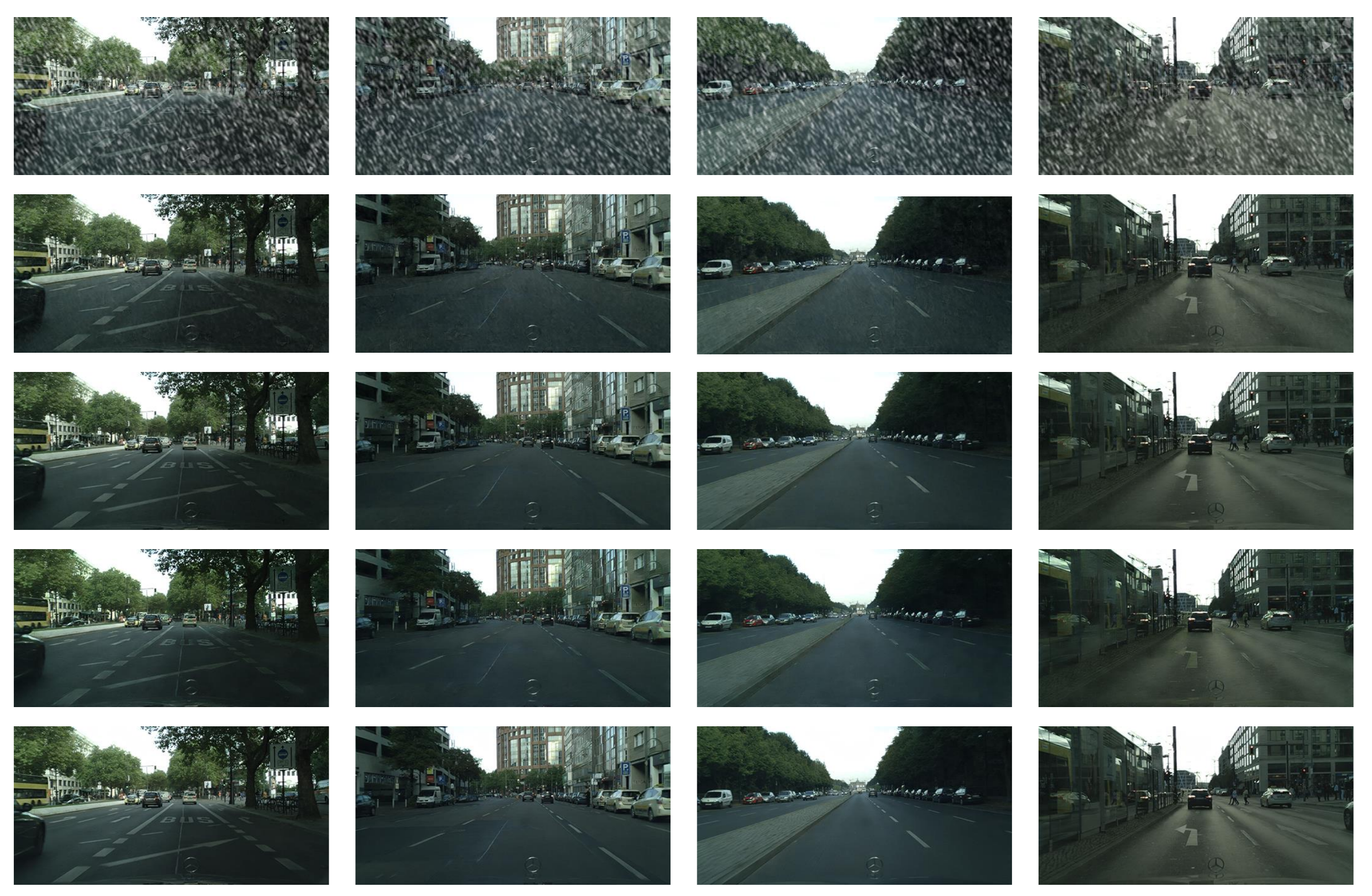}}
  \caption{{\bf Exemplar results on the SnowCityScapes dataset}. From top to bottom: input, DesnowNet, RESCAN, SPANet and Ours. Best viewed in color.}
  \label{fig:cityscapes} 
\end{figure*}

\item
\textbf{MSNet} is a multi-scale version of the above SnowCNN architecture. The main difference is that MSNet uses a multi-scale scheme to conduct the snow removal task like \cite{nah2017deep}. The input images are resized to different scales to help achieve finer results. The sub-networks in different scales share weights in our experiments.

\begin{figure*}[t] 
  \centering
  {\includegraphics[width=0.99\linewidth]{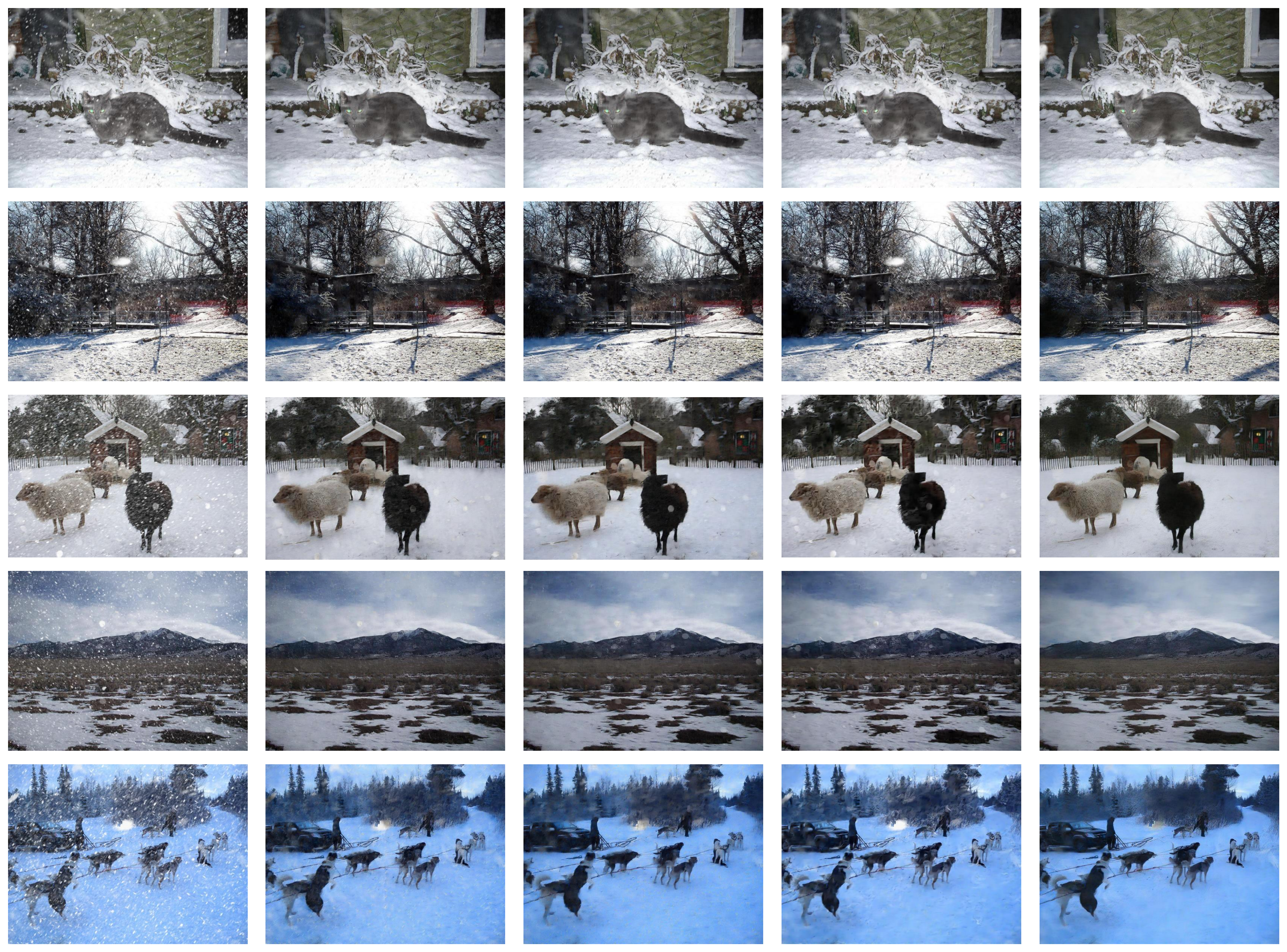}}
  \caption{{\bf Exemplar results on the Snow100K dataset}. From top to bottom: input, DesnowNet, RESCAN, SPANet and Ours. Best viewed in color.}
  \label{fig:snow100k} 
\end{figure*}

\item
\textbf{DDMSNet} is a dense multi-scale version of the above MSNet architecture. The main difference is that this model not only extracts features from images in different scales, but also extracts multi-scale features from images in a fixed scale.

\item
\textbf{DDMSNet(+)} is a coarse-to-fine version of the above DDMSNet architecture. We first use the SnowCNN to generate coarsely desnowed images and then feed them into the DDMSNet model to obtain finer results.

\item
\textbf{DDMSNet(S)} and \textbf{DDMSNet(G)} are two variants of the above DDMSNet(+). They majorly differ from the DDMSNet(+) due to that we use semantic-aware and geometry-aware representations to help update the DDMSNet, respectively.

\begin{figure*}[t] 
  \centering
  {\includegraphics[width=0.99\linewidth]{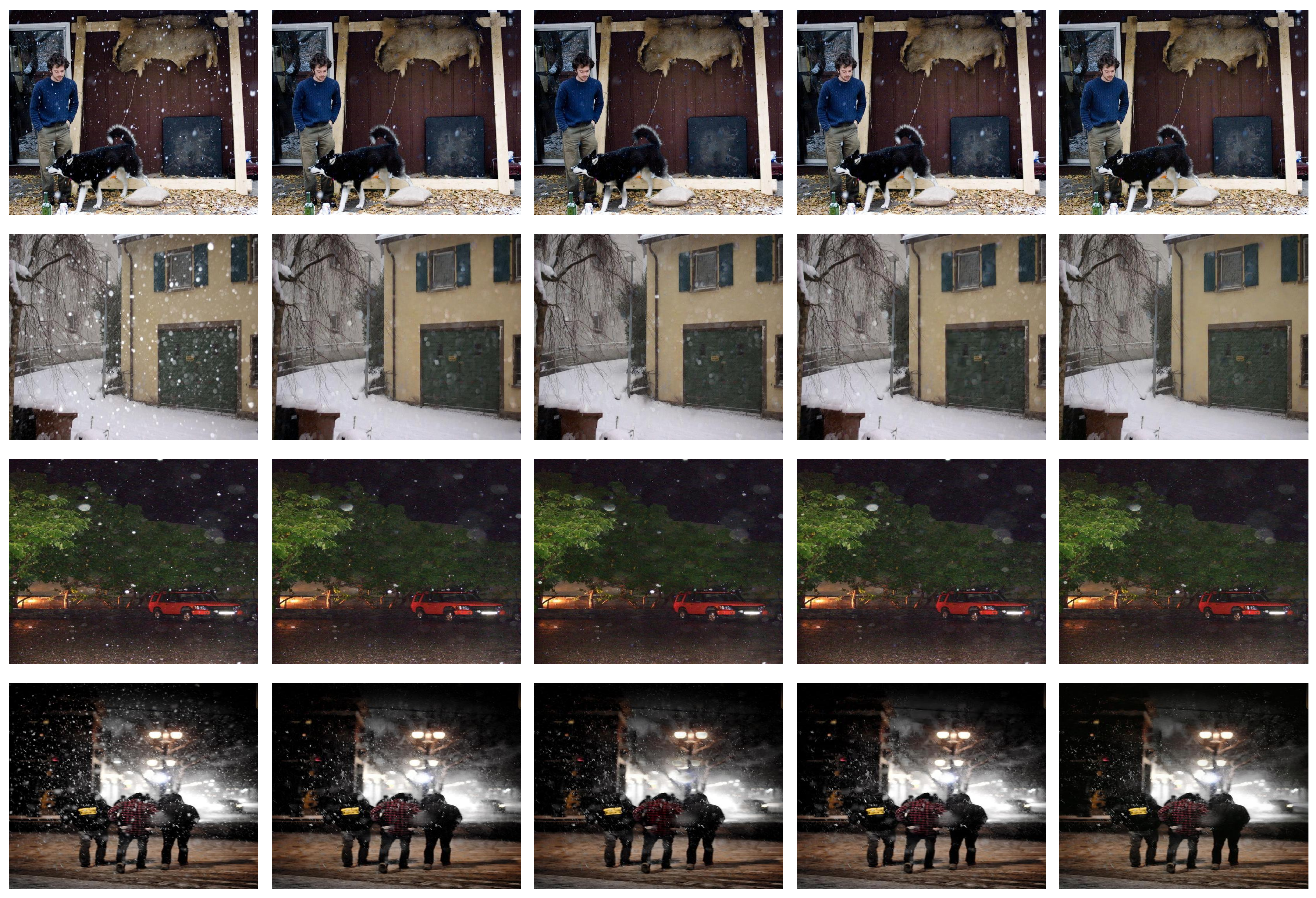}}
  \caption{{\bf Exemplar results on the real-world snowy frames}. From left to right: input, DesnowNet, RESCAN, SPANet and Ours. Best viewed in color.}
  \label{fig:real} 
\end{figure*}

\item
\textbf{DDMSNet(D+S)} is our final model. The snowy images are fed into a SnowCNN to obtain the coarse results, which are fed into the DDMSNet to learn the semantic-aware and geometry-aware representations to help restore the final clean images.

\end{itemize}

Fig. \ref{fig:ablation} and Table \ref{table:ablation} show the ablation study results of different variants on the SnowKITTI2012 dataset. In general, both the plain SnowCNN and MSNet achieve reasonable performance, but the performance is inferior when compared with the variants of our DDMSNet. The DDMSNet(+) performs better than the original DDMSNet, suggesting the effectiveness of the ``coarse-to-fine" strategy. With additional semantic-aware and the geometry-aware representation, the DDMSNet(S) and DDMSNet(G) respectively obtain better results compared with the variant DDMSNet(+). This verifies the usefulness of the introduced semantic-aware and geometry-aware attentions in the fine snow removal network. Finally, the DDMSNet(S+G) achieves the best performance without doubt, indicating the effect of the combination of semantic-aware and geometry-aware representations. 

It is also notable that these variants perform best in the case of small particle size, i.e., light snow. And in general they perform worse when the snow becomes heavier, though with a few inconsistent cases. This is reasonable and consistent with human perception.

\subsection{Comparison with Existing Methods}
\label{comparison}

To verify our model, we compare it with the existing state-of-the-art methods on the three datasets described above. To the best of our knowledge, there seems difficult to find existing methods specifically for snow removal, except the \textit{DesnowNet} \cite{liu2018desnownet}. To make the comparison more convincing, we additionally employ two learning-based rain removal approaches, \textit{RESCAN} \cite{li2018recurrent} and \textit{SPANet} \cite{wang2019spatial}, for the comparison. To adopt to the snow scenery, we retrain their networks with the snow dataset in the comparison.

Table \ref{table:sota} presents the quantitative results of different methods on the three datasets of SnowKITTI2012, SnowCityScapes and Snow100K. It shows that, the RESCAN and SPANet perform better than DesnowNet on the datasets of SnowKITTI2012 and SnowCityScapes, while worse on the Snow100K dataset. On all the three datasets, the performance achieved by our proposed method is significantly better than that of the counterparts.

Fig. \ref{fig:kitti}, Fig. \ref{fig:cityscapes} and Fig. \ref{fig:snow100k} represent the visual comparison results on the three datasets of SnowKITTI2012, SnowCityScapes and Snow100K. Compared with RESCAN, SPANet and DesnowNet, the results of our method exhibit fewer artifacts. And we achieve the best visually appealing results for snow removal.

\subsection{Performance in Real-World Scenarios}
\label{real_world}

In order to further verify the effectiveness of the proposed model in the real-world scenery, we compare the performance of our method with current state-of-the-art methods on real-world snowy images from the dataset of Snow100K. The qualitative results are shown in Fig. \ref{fig:real}, which validate that our method outperforms the current methods on real-world snowy images.

\section{Conclusion}

We propose a new multi-scale network, named as Deep Dense Multi-scale Network (\textit{DDMSNet}), and demonstrate its superior performance for snow removal. We exploit the semantic and geometric information as global priors to better remove snow and restore the clean images. Furthermore, based on the public KITTI and Cityscapes datasets, we synthesize two large-scale snowy datasets for snow removal. Experimental results demonstrate that the proposed method performs better than previous methods and achieves state-of-the-art performance.

\section*{Acknowledgment}
This work is funded in part by the ARC Centre of Excellence for Robotics Vision (CE140100016),  ARC-Discovery (DP 190102261) and ARC-LIEF (190100080) grants, as well as a research grant from Baidu on autonomous driving.  The authors gratefully acknowledge the GPUs donated by NVIDIA Corporation. We thank all anonymous reviewers and editors for their constructive comments.


\bibliographystyle{IEEEtran}
\bibliography{egbib}

\begin{thebibliography}{10}
\providecommand{\url}[1]{#1}
\csname url@samestyle\endcsname
\providecommand{\newblock}{\relax}
\providecommand{\bibinfo}[2]{#2}
\providecommand{\BIBentrySTDinterwordspacing}{\spaceskip=0pt\relax}
\providecommand{\BIBentryALTinterwordstretchfactor}{4}
\providecommand{\BIBentryALTinterwordspacing}{\spaceskip=\fontdimen2\font plus
\BIBentryALTinterwordstretchfactor\fontdimen3\font minus
  \fontdimen4\font\relax}
\providecommand{\BIBforeignlanguage}[2]{{%
\expandafter\ifx\csname l@#1\endcsname\relax
\typeout{** WARNING: IEEEtran.bst: No hyphenation pattern has been}%
\typeout{** loaded for the language `#1'. Using the pattern for}%
\typeout{** the default language instead.}%
\else
\language=\csname l@#1\endcsname
\fi
#2}}
\providecommand{\BIBdecl}{\relax}
\BIBdecl

\bibitem{dalal2005histograms}
N.~Dalal and B.~Triggs, ``Histograms of oriented gradients for human
  detection,'' in \emph{2005 IEEE computer society conference on computer
  vision and pattern recognition (CVPR'05)}, vol.~1.\hskip 1em plus 0.5em minus
  0.4em\relax IEEE, 2005, pp. 886--893.

\bibitem{kaempfer2007three}
T.~U. Kaempfer, M.~Hopkins, and D.~Perovich, ``A three-dimensional
  microstructure-based photon-tracking model of radiative transfer in snow,''
  \emph{Journal of Geophysical Research: Atmospheres}, vol. 112, no. D24, 2007.

\bibitem{itti1998model}
L.~Itti, C.~Koch, and E.~Niebur, ``A model of saliency-based visual attention
  for rapid scene analysis,'' \emph{IEEE Transactions on pattern analysis and
  machine intelligence}, vol.~20, no.~11, pp. 1254--1259, 1998.

\bibitem{liu2018desnownet}
Y.-F. Liu, D.-W. Jaw, S.-C. Huang, and J.-N. Hwang, ``Desnownet: Context-aware
  deep network for snow removal,'' \emph{IEEE Transactions on Image
  Processing}, vol.~27, no.~6, pp. 3064--3073, 2018.

\bibitem{fourure2017residual}
D.~Fourure, R.~Emonet, E.~Fromont, D.~Muselet, A.~Tremeau, and C.~Wolf,
  ``Residual conv-deconv grid network for semantic segmentation,'' \emph{arXiv
  preprint arXiv:1707.07958}, 2017.

\bibitem{wang2017hierarchical}
Y.~Wang, S.~Liu, C.~Chen, and B.~Zeng, ``A hierarchical approach for rain or
  snow removing in a single color image,'' \emph{IEEE Transactions on Image
  Processing}, vol.~26, no.~8, pp. 3936--3950, 2017.

\bibitem{bossu2011rain}
J.~Bossu, N.~Hauti{\`e}re, and J.-P. Tarel, ``Rain or snow detection in image
  sequences through use of a histogram of orientation of streaks,''
  \emph{International journal of computer vision}, vol.~93, no.~3, pp.
  348--367, 2011.

\bibitem{pei2014removing}
S.-C. Pei, Y.-T. Tsai, and C.-Y. Lee, ``Removing rain and snow in a single
  image using saturation and visibility features,'' in \emph{2014 IEEE
  International Conference on Multimedia and Expo Workshops (ICMEW)}.\hskip 1em
  plus 0.5em minus 0.4em\relax IEEE, 2014, pp. 1--6.

\bibitem{rajderkar2013removing}
D.~Rajderkar and P.~Mohod, ``Removing snow from an image via image
  decomposition,'' in \emph{2013 IEEE International Conference ON Emerging
  Trends in Computing, Communication and Nanotechnology (ICECCN)}.\hskip 1em
  plus 0.5em minus 0.4em\relax IEEE, 2013, pp. 576--579.

\bibitem{xu2012improved}
J.~Xu, W.~Zhao, P.~Liu, and X.~Tang, ``An improved guidance image based method
  to remove rain and snow in a single image,'' \emph{Computer and Information
  Science}, vol.~5, no.~3, p.~49, 2012.

\bibitem{xu2012removing}
------, ``Removing rain and snow in a single image using guided filter,'' in
  \emph{2012 IEEE International Conference on Computer Science and Automation
  Engineering (CSAE)}, vol.~2.\hskip 1em plus 0.5em minus 0.4em\relax IEEE,
  2012, pp. 304--307.

\bibitem{ledig2017photo}
C.~Ledig, L.~Theis, F.~Husz{\'a}r, J.~Caballero, A.~Cunningham, A.~Acosta,
  A.~Aitken, A.~Tejani, J.~Totz, Z.~Wang \emph{et~al.}, ``Photo-realistic
  single image super-resolution using a generative adversarial network,'' in
  \emph{Proceedings of the IEEE Conference on Computer Vision and Pattern
  Recognition (CVPR)}, 2017.

\bibitem{johnson2016perceptual}
J.~Johnson, A.~Alahi, and L.~Fei-Fei, ``Perceptual losses for real-time style
  transfer and super-resolution,'' in \emph{European Conference on Computer
  Vision (ECCV)}, 2016.

\bibitem{zhang2018adversarial}
K.~Zhang, W.~Luo, Y.~Zhong, L.~Ma, W.~Liu, and H.~Li, ``Adversarial
  spatio-temporal learning for video deblurring,'' \emph{IEEE Transactions on
  Image Processing (TIP)}, 2018.

\bibitem{zhang2020deblurring}
K.~Zhang, W.~Luo, Y.~Zhong, L.~Ma, B.~Stenger, W.~Liu, and H.~Li, ``Deblurring
  by realistic blurring,'' in \emph{Proceedings of the IEEE Conference on
  Computer Vision and Pattern Recognition (CVPR)}, 2020.

\bibitem{li2021arvo}
D.~Li, C.~Xu, K.~Zhang, X.~Yu, Y.~Zhong, W.~Ren, H.~Suominen, and H.~Li,
  ``Arvo: Learning all-range volumetric correspondence for video deblurring,''
  \emph{arXiv preprint arXiv:2103.04260}, 2021.

\bibitem{zhang2020every}
K.~Zhang, W.~Luo, B.~Stenger, W.~Ren, L.~Ma, and H.~Li, ``Every moment matters:
  Detail-aware networks to bring a blurry image alive,'' in \emph{Proceedings
  of the 28th ACM International Conference on Multimedia}, 2020, pp. 384--392.

\bibitem{zhang2021dual}
K.~Zhang, D.~Li, W.~Luo, W.~Ren, L.~Ma, and H.~Li, ``Dual
  attention-in-attention model for joint rain streak and raindrop removal,''
  \emph{arXiv preprint arXiv:2103.07051}, 2021.

\bibitem{zhang2020beyond}
K.~Zhang, W.~Luo, W.~Ren, J.~Wang, F.~Zhao, L.~Ma, and H.~Li, ``Beyond
  monocular deraining: Stereo image deraining via semantic understanding,'' in
  \emph{European Conference on Computer Vision}.\hskip 1em plus 0.5em minus
  0.4em\relax Springer, 2020, pp. 71--89.

\bibitem{li2019single}
S.~Li, I.~B. Araujo, W.~Ren, Z.~Wang, E.~K. Tokuda, R.~H. Junior,
  R.~Cesar-Junior, J.~Zhang, X.~Guo, and X.~Cao, ``Single image deraining: A
  comprehensive benchmark analysis,'' in \emph{Proceedings of the IEEE
  Conference on Computer Vision and Pattern Recognition (CVPR)}, 2019.

\bibitem{li2019stacked}
P.~Li, M.~Yun, J.~Tian, Y.~Tang, G.~Wang, and C.~Wu, ``Stacked dense networks
  for single-image snow removal,'' \emph{Neurocomputing}, vol. 367, pp.
  152--163, 2019.

\bibitem{li2020all}
R.~Li, R.~T. Tan, and L.-F. Cheong, ``All in one bad weather removal using
  architectural search,'' in \emph{Proceedings of the IEEE/CVF Conference on
  Computer Vision and Pattern Recognition}, 2020, pp. 3175--3185.

\bibitem{kang2011automatic}
L.-W. Kang, C.-W. Lin, and Y.-H. Fu, ``Automatic single-image-based rain
  streaks removal via image decomposition,'' \emph{IEEE transactions on image
  processing}, vol.~21, no.~4, pp. 1742--1755, 2011.

\bibitem{luo2015removing}
Y.~Luo, Y.~Xu, and H.~Ji, ``Removing rain from a single image via
  discriminative sparse coding,'' in \emph{Proceedings of the IEEE
  International Conference on Computer Vision}, 2015, pp. 3397--3405.

\bibitem{li2016rain}
Y.~Li, R.~T. Tan, X.~Guo, J.~Lu, and M.~S. Brown, ``Rain streak removal using
  layer priors,'' in \emph{Proceedings of the IEEE conference on computer
  vision and pattern recognition}, 2016, pp. 2736--2744.

\bibitem{chang2017transformed}
Y.~Chang, L.~Yan, and S.~Zhong, ``Transformed low-rank model for line pattern
  noise removal,'' in \emph{Proceedings of the IEEE International Conference on
  Computer Vision}, 2017, pp. 1726--1734.

\bibitem{zhu2017joint}
L.~Zhu, C.-W. Fu, D.~Lischinski, and P.-A. Heng, ``Joint bi-layer optimization
  for single-image rain streak removal,'' in \emph{Proceedings of the IEEE
  international conference on computer vision}, 2017, pp. 2526--2534.

\bibitem{du2018single}
S.~Du, Y.~Liu, M.~Ye, Z.~Xu, J.~Li, and J.~Liu, ``Single image deraining via
  decorrelating the rain streaks and background scene in gradient domain,''
  \emph{Pattern Recognition}, vol.~79, pp. 303--317, 2018.

\bibitem{chen2014visual}
D.-Y. Chen, C.-C. Chen, and L.-W. Kang, ``Visual depth guided color image rain
  streaks removal using sparse coding,'' \emph{IEEE transactions on circuits
  and systems for video technology}, vol.~24, no.~8, pp. 1430--1455, 2014.

\bibitem{zhang2019image}
H.~Zhang, V.~Sindagi, and V.~M. Patel, ``Image de-raining using a conditional
  generative adversarial network,'' \emph{IEEE Transactions on Circuits and
  Systems for Video Technology (TCSVT)}, 2019.

\bibitem{fu2017clearing}
X.~Fu, J.~Huang, X.~Ding, Y.~Liao, and J.~Paisley, ``Clearing the skies: A deep
  network architecture for single-image rain removal,'' \emph{IEEE Transactions
  on Image Processing (TIP)}, 2017.

\bibitem{fu2017removing}
X.~Fu, J.~Huang, D.~Zeng, Y.~Huang, X.~Ding, and J.~Paisley, ``Removing rain
  from single images via a deep detail network,'' in \emph{Proceedings of the
  IEEE Conference on Computer Vision and Pattern Recognition (CVPR)}, 2017.

\bibitem{yang2017deep}
W.~Yang, R.~T. Tan, J.~Feng, J.~Liu, Z.~Guo, and S.~Yan, ``Deep joint rain
  detection and removal from a single image,'' in \emph{Proceedings of the IEEE
  Conference on Computer Vision and Pattern Recognition (CVPR)}, 2017.

\bibitem{zhang2018density}
H.~Zhang and V.~M. Patel, ``Density-aware single image de-raining using a
  multi-stream dense network,'' in \emph{Proceedings of the IEEE Conference on
  Computer Vision and Pattern Recognition (CVPR)}, 2018.

\bibitem{li2018recurrent}
X.~Li, J.~Wu, Z.~Lin, H.~Liu, and H.~Zha, ``Recurrent squeeze-and-excitation
  context aggregation net for single image deraining,'' in \emph{European
  Conference on Computer Vision (ECCV)}, 2018.

\bibitem{eigen2013restoring}
D.~Eigen, D.~Krishnan, and R.~Fergus, ``Restoring an image taken through a
  window covered with dirt or rain,'' in \emph{Proceedings of the IEEE
  International Conference on Computer Vision (ICCV)}, 2013.

\bibitem{qian2018attentive}
R.~Qian, R.~T. Tan, W.~Yang, J.~Su, and J.~Liu, ``Attentive generative
  adversarial network for raindrop removal from a single image,'' in
  \emph{Proceedings of the IEEE Conference on Computer Vision and Pattern
  Recognition (CVPR)}, 2018.

\bibitem{zheng2019residual}
Y.~Zheng, X.~Yu, M.~Liu, and S.~Zhang, ``Residual multiscale based single image
  deraining.'' in \emph{British Machine Vision Conference (BMVC)}, 2019.

\bibitem{li2019heavy}
R.~Li, L.-F. Cheong, and R.~T. Tan, ``Heavy rain image restoration: Integrating
  physics model and conditional adversarial learning,'' in \emph{Proceedings of
  the IEEE Conference on Computer Vision and Pattern Recognition}, 2019, pp.
  1633--1642.

\bibitem{schechner2001instant}
Y.~Y. Schechner, S.~G. Narasimhan, and S.~K. Nayar, ``Instant dehazing of
  images using polarization,'' in \emph{Proceedings of the 2001 IEEE Computer
  Society Conference on Computer Vision and Pattern Recognition. CVPR 2001},
  vol.~1.\hskip 1em plus 0.5em minus 0.4em\relax IEEE, 2001, pp. I--I.

\bibitem{shwartz2006blind}
S.~Shwartz, E.~Namer, and Y.~Y. Schechner, ``Blind haze separation,'' in
  \emph{2006 IEEE Computer Society Conference on Computer Vision and Pattern
  Recognition (CVPR'06)}, vol.~2.\hskip 1em plus 0.5em minus 0.4em\relax IEEE,
  2006, pp. 1984--1991.

\bibitem{narasimhan2000chromatic}
S.~G. Narasimhan and S.~K. Nayar, ``Chromatic framework for vision in bad
  weather,'' in \emph{Proceedings IEEE Conference on Computer Vision and
  Pattern Recognition. CVPR 2000 (Cat. No. PR00662)}, vol.~1.\hskip 1em plus
  0.5em minus 0.4em\relax IEEE, 2000, pp. 598--605.

\bibitem{narasimhan2003contrast}
------, ``Contrast restoration of weather degraded images,'' \emph{IEEE
  transactions on pattern analysis and machine intelligence}, vol.~25, no.~6,
  pp. 713--724, 2003.

\bibitem{nayar1999vision}
S.~K. Nayar and S.~G. Narasimhan, ``Vision in bad weather,'' in
  \emph{Proceedings of the Seventh IEEE International Conference on Computer
  Vision}, vol.~2.\hskip 1em plus 0.5em minus 0.4em\relax IEEE, 1999, pp.
  820--827.

\bibitem{cozman1997depth}
F.~Cozman and E.~Krotkov, ``Depth from scattering,'' in \emph{Proceedings of
  IEEE Computer Society Conference on Computer Vision and Pattern
  Recognition}.\hskip 1em plus 0.5em minus 0.4em\relax IEEE, 1997, pp.
  801--806.

\bibitem{narasimhan2002vision}
S.~G. Narasimhan and S.~K. Nayar, ``Vision and the atmosphere,''
  \emph{International journal of computer vision}, vol.~48, no.~3, pp.
  233--254, 2002.

\bibitem{kopf2008deep}
J.~Kopf, B.~Neubert, B.~Chen, M.~Cohen, D.~Cohen-Or, O.~Deussen,
  M.~Uyttendaele, and D.~Lischinski, ``Deep photo: Model-based photograph
  enhancement and viewing,'' \emph{ACM transactions on graphics (TOG)},
  vol.~27, no.~5, pp. 1--10, 2008.

\bibitem{tarel2009fast}
J.-P. Tarel and N.~Hautiere, ``Fast visibility restoration from a single color
  or gray level image,'' in \emph{2009 IEEE 12th International Conference on
  Computer Vision}.\hskip 1em plus 0.5em minus 0.4em\relax IEEE, 2009, pp.
  2201--2208.

\bibitem{he2010single}
K.~He, J.~Sun, and X.~Tang, ``Single image haze removal using dark channel
  prior,'' \emph{IEEE transactions on pattern analysis and machine
  intelligence}, vol.~33, no.~12, pp. 2341--2353, 2010.

\bibitem{meng2013efficient}
G.~Meng, Y.~Wang, J.~Duan, S.~Xiang, and C.~Pan, ``Efficient image dehazing
  with boundary constraint and contextual regularization,'' in
  \emph{Proceedings of the IEEE international conference on computer vision},
  2013, pp. 617--624.

\bibitem{li2015nighttime}
Y.~Li, R.~T. Tan, and M.~S. Brown, ``Nighttime haze removal with glow and
  multiple light colors,'' in \emph{Proceedings of the IEEE international
  conference on computer vision}, 2015, pp. 226--234.

\bibitem{nishino2012bayesian}
K.~Nishino, L.~Kratz, and S.~Lombardi, ``Bayesian defogging,''
  \emph{International journal of computer vision}, vol.~98, no.~3, pp.
  263--278, 2012.

\bibitem{fattal2014dehazing}
R.~Fattal, ``Dehazing using color-lines,'' \emph{ACM transactions on graphics
  (TOG)}, vol.~34, no.~1, pp. 1--14, 2014.

\bibitem{berman2016non}
D.~Berman, S.~Avidan \emph{et~al.}, ``Non-local image dehazing,'' in
  \emph{Proceedings of the IEEE conference on computer vision and pattern
  recognition}, 2016, pp. 1674--1682.

\bibitem{cai2016dehazenet}
B.~Cai, X.~Xu, K.~Jia, C.~Qing, and D.~Tao, ``Dehazenet: An end-to-end system
  for single image haze removal,'' \emph{IEEE Transactions on Image
  Processing}, vol.~25, no.~11, pp. 5187--5198, 2016.

\bibitem{ren2016single}
W.~Ren, S.~Liu, H.~Zhang, J.~Pan, X.~Cao, and M.-H. Yang, ``Single image
  dehazing via multi-scale convolutional neural networks,'' in \emph{European
  Conference on Computer Vision (ECCV)}, 2016.

\bibitem{zhang2018densely}
H.~Zhang and V.~M. Patel, ``Densely connected pyramid dehazing network,'' in
  \emph{Proceedings of the IEEE conference on computer vision and pattern
  recognition}, 2018, pp. 3194--3203.

\bibitem{li2017aod}
B.~Li, X.~Peng, Z.~Wang, J.~Xu, and D.~Feng, ``Aod-net: All-in-one dehazing
  network,'' in \emph{Proceedings of the IEEE International Conference on
  Computer Vision (ICCV)}, 2017.

\bibitem{cheng2018semantic}
Z.~Cheng, S.~You, V.~Ila, and H.~Li, ``Semantic single-image dehazing,''
  \emph{arXiv preprint arXiv:1804.05624}, 2018.

\bibitem{huang2017densely}
G.~Huang, Z.~Liu, L.~Van Der~Maaten, and K.~Q. Weinberger, ``Densely connected
  convolutional networks,'' in \emph{Proceedings of the IEEE conference on
  computer vision and pattern recognition}, 2017, pp. 4700--4708.

\bibitem{tao2020hierarchical}
A.~Tao, K.~Sapra, and B.~Catanzaro, ``Hierarchical multi-scale attention for
  semantic segmentation,'' \emph{arXiv preprint arXiv:2005.10821}, 2020.

\bibitem{yin2019enforcing}
W.~Yin, Y.~Liu, C.~Shen, and Y.~Yan, ``Enforcing geometric constraints of
  virtual normal for depth prediction,'' in \emph{Proceedings of the IEEE
  International Conference on Computer Vision}, 2019, pp. 5684--5693.

\bibitem{wang2018esrgan}
X.~Wang, K.~Yu, S.~Wu, J.~Gu, Y.~Liu, C.~Dong, Y.~Qiao, and C.~Change~Loy,
  ``Esrgan: Enhanced super-resolution generative adversarial networks,'' in
  \emph{Proceedings of the European Conference on Computer Vision (ECCV)},
  2018, pp. 0--0.

\bibitem{simonyan2014very}
K.~Simonyan and A.~Zisserman, ``Very deep convolutional networks for
  large-scale image recognition,'' \emph{arXiv preprint arXiv:1409.1556}, 2014.

\bibitem{geiger2013vision}
A.~Geiger, P.~Lenz, C.~Stiller, and R.~Urtasun, ``Vision meets robotics: The
  kitti dataset,'' \emph{The International Journal of Robotics Research
  (IJRR)}, 2013.

\bibitem{cordts2016cityscapes}
M.~Cordts, M.~Omran, S.~Ramos, T.~Rehfeld, M.~Enzweiler, R.~Benenson,
  U.~Franke, S.~Roth, and B.~Schiele, ``The cityscapes dataset for semantic
  urban scene understanding,'' in \emph{Proceedings of the IEEE conference on
  computer vision and pattern recognition}, 2016, pp. 3213--3223.

\bibitem{nah2017deep}
S.~Nah, T.~Hyun~Kim, and K.~Mu~Lee, ``Deep multi-scale convolutional neural
  network for dynamic scene deblurring,'' in \emph{Proceedings of the IEEE
  Conference on Computer Vision and Pattern Recognition}, 2017, pp. 3883--3891.

\bibitem{wang2019spatial}
T.~Wang, X.~Yang, K.~Xu, S.~Chen, Q.~Zhang, and R.~W. Lau, ``Spatial attentive
  single-image deraining with a high quality real rain dataset,'' in
  \emph{Proceedings of the IEEE Conference on Computer Vision and Pattern
  Recognition}, 2019, pp. 12\,270--12\,279.

\end{thebibliography}
\end{document}